\documentclass[10pt, a4paper]{article}
\usepackage{lrec2026}
\usepackage{multibib}
\newcites{languageresource}{Language Resources}
\usepackage{graphicx}
\usepackage{tabularx}
\usepackage{soul}
\usepackage{hyperref}
\usepackage{latexsym}
\usepackage{booktabs}
\usepackage{multirow}
\usepackage{amsmath}
\usepackage[utf8]{inputenc}
\usepackage[T1]{fontenc}
\usepackage[arabic, english]{babel}
\usepackage{times}

\title{QU-NLP at QIAS 2026: Multi-Stage QLoRA Fine-Tuning for Arabic Islamic Inheritance Reasoning}

\name{Mohammad AL-Smadi}

\address{Qatar University \\
         Doha, Qatar \\
         malsmadi@qu.edu.qa}

\abstract{
Islamic inheritance law (\textAR{علم المواريث}, \textit{'ilm al-maw\=ar\={\i}th}) presents a challenging domain for evaluating large language models' structured reasoning capabilities, requiring multi-step legal analysis, rule-based blocking decisions, and precise fractional calculations. We present QU-NLP's submission to the QIAS 2026 shared task on Arabic Islamic inheritance reasoning. Our approach employs a multi-stage Quantized Low-Rank Adaptation (QLoRA) fine-tuning strategy on Qwen3-4B: (1) domain adaptation on 3,166 Islamic fatwa records to acquire inheritance terminology and jurisprudential reasoning patterns, followed by (2) task-specific training on 12,000 structured inheritance cases to optimize JSON-formatted output generation. Using 4-bit NF4 quantization with rank-128 LoRA adapters, our model achieves 90\% MIR-E (Mawarith Inheritance Reasoning Evaluation) score on the test set, demonstrating competitive performance while requiring minimal computational resources. Our results show that domain-specific pre-adaptation combined with structured output training enables small language models to perform complex legal reasoning tasks effectively, matching commercial systems such as Gemini-2.5-flash. 
}

\begin{document}

\maketitleabstract

\section{Introduction}

Large language models (LLMs) have demonstrated remarkable capabilities across diverse natural language processing tasks~\cite{openai2024gpt4}. However, their ability to perform structured, rule-based reasoning under strict legal constraints remains insufficiently evaluated. Islamic inheritance law (\textAR{علم المواريث}, \textit{'ilm al-maw\=ar\={\i}th}) offers a particularly demanding testbed for evaluating multi-step legal reasoning capabilities~\cite{bouchekif2025mawarith}.

Solving an Islamic inheritance case requires a well-defined procedural chain: (1) identifying eligible heirs from a textual description of family relations, (2) applying blocking rules (\textAR{حجب}, \textit{\d{h}ajb}) to determine which relatives are excluded by closer heirs, (3) assigning prescribed Qur'anic shares to eligible heirs, (4) detecting and applying adjustment mechanisms such as \textAR{عول} (\textit{'awl}, proportional reduction when shares exceed unity) or \textAR{رد} (\textit{radd}, redistribution of surplus), and (5) computing the final normalized distribution. Errors at any intermediate stage propagate deterministically and invalidate subsequent calculations, making this domain particularly suitable for evaluating structured reasoning under jurisprudential constraints.

The QIAS 2026 shared task represents a significant evolution in evaluating models on Islamic inheritance reasoning. While the 2025 task assessed models through multiple-choice questions~\cite{bouchekif2025qias}, the 2026 task introduces MAWARITH, a large-scale dataset of 12,500 Arabic inheritance cases with detailed step-by-step reasoning annotations~\cite{bouchekif2025mawarith}. QIAS 2026 requires generating complete structured reasoning traces in JSON format, exposing all intermediate legal decisions across a five-stage pipeline: heir identification, blocking rule application, share calculation, adjustment detection, and final distribution. This methodological shift addresses critical limitations of MCQ evaluation where models can succeed through memorization without genuine understanding, and where binary scoring provides no diagnostic insights into specific failure modes. The structured output requirement enables the multi-component MIR-E evaluation metric, which assigns partial credit for correct intermediate steps---for instance, awarding up to 70\% to cases with perfect legal reasoning but arithmetic errors. This evaluation framework enables our fine-grained error analysis identifying four distinct categories of model failures with targeted improvement strategies, analysis impossible under answer-selection formats.

We present QU-NLP's approach to this challenge, employing a multi-stage QLoRA fine-tuning strategy on Qwen3-4B~\cite{Qwen2025}. Our key contributions are:

\begin{itemize}
\item A two-stage training methodology combining domain adaptation on Islamic legal texts with task-specific fine-tuning on structured inheritance solutions.
\item Demonstration that 4-bit quantized models with LoRA adapters can achieve a MIR-E score of 90\% on complex multi-step legal reasoning while requiring minimal computational resources, placing it among the top-performing systems and significantly outperforming larger open-weight models evaluated in the baseline study~\cite{bouchekif2025assessing}.
\item Comprehensive error analysis identifying four distinct failure modes and their root causes, providing actionable insights for model improvement.
\end{itemize}

\section{Related Work}

\subsection{LLMs for Islamic Knowledge Tasks}

Recent work has explored LLMs for Islamic knowledge tasks including Qur'anic question answering~\cite{bhatia2026rag,malhas2022quran} and hallucination detection in Islamic content~\cite{mubarak2025islamiceval}. These studies reveal that while LLMs perform adequately on retrieval-based tasks relying on textual matching, they exhibit significant limitations on tasks requiring structured reasoning or deep domain knowledge. Bouchekif et al.~\cite{bouchekif2025assessing} assess LLMs on Islamic legal reasoning, identifying systematic failures in inheritance case resolution and raising concerns about model reliability in religious and legal applications.

Retrieval-Augmented Generation (RAG) approaches have been explored to improve answer grounding~\cite{alsmadi2025qunlp2025,alowaidi2025sea}. However, RAG remains insufficient for questions requiring multi-step inference, motivating the development of reasoning-oriented models.

Within Islamic inheritance specifically, prior work has focused on multiple-choice evaluations. QIAS 2025~\cite{bouchekif2025qias} introduced a shared task on Islamic inheritance reasoning assessed through MCQs, where models select correct answers without exposing reasoning traces. Elrefai et al.~\cite{elrefai2025gumball} participated in QIAS 2025 with a fine-tuned Arabic LLM, but the MCQ format prevented assessment of whether models truly reason correctly or merely pattern-match. AL-Smadi~\cite{alsmadi2025qunlp2025} explored a two-phase fine-tuning approach combined with Retrieval-Augmented Generation at QIAS~2025, investigating hybrid retrieval and generation strategies for Islamic inheritance reasoning. MirathQA~\cite{almasoud2026mirathqa} provides a dataset of Hanbali inheritance cases in MCQ format.

The \textit{MAWARITH} dataset~\cite{bouchekif2025mawarith} addresses this limitation by requiring end-to-end reasoning generation with intermediate justifications, enabling fine-grained error analysis across the inheritance reasoning pipeline.

\subsection{Legal Reasoning with LLMs}

Beyond Islamic domains, legal reasoning benchmarks have emerged to evaluate LLMs on structured argumentation and rule-based inference. LegalBench~\cite{guha2023legalbench}, LexGLUE~\cite{chalkidis2022lexglue}, and LEXTREME~\cite{niklaus2023lextreme} assess models on common-law legal tasks. BRIEFME~\cite{woo2025briefme} evaluates legal argument summarization in the context of assisting with legal briefs.

Recent models explicitly designed for multi-step reasoning include GPT-5~\cite{singh2025openai}, Gemini~\cite{anil2023gemini}, DeepSeek-R1~\cite{deepseek2024r1}, and open-weight alternatives such as Qwen3~\cite{Qwen2025} and Fanar~\cite{abbas2025fanar}. These models promote consistent multi-step inference through instruction tuning and reinforcement learning.

Parameter-efficient fine-tuning methods, particularly QLoRA~\cite{dettmers2023qlora}, enable adaptation of large models with limited resources. QLoRA combines 4-bit NF4 quantization with Low-Rank Adaptation~\cite{hu2021lora}, achieving competitive fine-tuning performance while dramatically reducing memory requirements. This approach has been applied to domain adaptation in specialized Arabic NLP tasks~\cite{alsmadi2025qunlp2025}.

\section{Task and Dataset}

\subsection{Task Definition}

The QIAS 2026 shared task requires models to solve Arabic Islamic inheritance cases by generating structured JSON outputs that expose all intermediate reasoning steps. Given a natural-language description of the deceased and surviving relatives, models must execute a complete five-stage reasoning pipeline without access to gold intermediate steps.

\subsubsection{JSON Output Structure}

As explained in the \textit{MAWARITH} dataset~\cite{bouchekif2025mawarith}, the required JSON output contains five mandatory components representing distinct reasoning stages:

\textbf{1. Heirs} (\textAR{الورثة}, \textit{al-waratha}): A list identifying all eligible inheriting relatives with their counts. Each heir entry specifies the heir category (e.g., \textAR{ابن} ``son'', \textAR{بنت} ``daughter'', \textAR{أم} ``mother'') and the number of individuals in that category. This stage requires applying Qur'anic eligibility rules based on kinship relationships.

\textbf{2. Blocked} (\textAR{المحجوبون}, \textit{al-ma\d{h}j\=ub\=un}): A list of relatives mentioned in the scenario who are present but excluded from inheritance due to blocking rules (\textAR{حجب}, \textit{\d{h}ajb}). Islamic inheritance law stipulates that closer relatives block more distant ones in specific patterns---for instance, a son blocks grandsons, and a father blocks uncles. Correctly identifying blocked heirs demonstrates understanding of these hierarchical rules.

\textbf{3. Shares} (\textAR{الأنصبة}, \textit{al-an\d{s}iba}): Initial prescribed fractional shares assigned to each eligible heir before any global adjustments. The Qur'an specifies fixed shares for certain heir categories (e.g., wife receives 1/4 if no children, 1/8 if children exist; daughter receives 1/2 if alone, 2/3 if multiple). Residuary heirs (\textAR{عصبة}, \textit{'a\d{s}aba}) such as sons and brothers receive the remainder after fixed shares are distributed, designated as \textAR{باقى التركة} (``remainder of estate'') rather than numerical fractions.

\textbf{4. 'Awl or Radd} (\textAR{عول أو رد}): The type of global adjustment mechanism applied when prescribed shares do not sum to exactly the full estate:

\begin{itemize}
\item \textbf{\textAR{رد} (Radd, redistribution)}: Applies when total prescribed shares are less than the full estate \textit{and} no residuary heir is present. After distributing all fixed shares, the remaining unassigned portion is redistributed proportionally among eligible fixed-share heirs according to their original shares. For example, if a mother receives 1/6 and a daughter receives 1/2, the total is 1/6 + 1/2 = 2/3, leaving 1/3 unassigned. This remainder is redistributed through \textit{radd}, increasing each heir's allocation proportionally to their original prescribed shares.

\item \textbf{\textAR{عول} ('Awl, proportional reduction)}: Applies when total prescribed shares exceed the full estate. Since the initially assigned shares cannot all be satisfied in full, all shares are proportionally reduced so their sum equals exactly the estate. For instance, if prescribed shares sum to 1/2 + 1/6 + 2/3 = 8/6 (greater than 1), each share is scaled down through \textit{'awl} so that the total distribution fits within the estate.

\item \textbf{\textAR{لا} (None)}: No adjustment needed when shares sum to exactly the estate or when residuary heirs absorb the remainder naturally.
\end{itemize}

\textbf{5. Post-Tasil} (\textAR{بعد التأصيل}, \textit{ba'da al-ta\d{s}\={\i}l}): The final normalized distribution after applying any adjustments. This component contains:
\begin{itemize}
\item \texttt{total\_shares}: The denominator of the fractional distribution after adjustment.
\item \texttt{distribution}: A list specifying each heir's final allocation as both fractional shares (e.g., ``3/12'') and normalized percentages (e.g., 25.0\%).
\end{itemize}

This stage requires precise numerical computation to ensure all percentages sum to exactly 100\% and correctly reflect the applied adjustments.

\subsubsection{Task Complexity}

While Islamic inheritance follows deterministic jurisprudential rules that could be implemented in a symbolic rule engine with perfect accuracy, the QIAS 2026 task evaluates a fundamentally different capability: \textit{end-to-end neural reasoning from natural language to structured output}. Unlike rule-based systems that operate on pre-structured inputs and apply explicit programmed logic, neural models must simultaneously solve multiple interdependent challenges:

\begin{itemize}
\item \textbf{Natural language understanding}: Models must parse Arabic text with diverse linguistic expressions to extract family relationships. The same heir category can be expressed through multiple lexical variants (e.g., \textAR{أم} ``mother'' vs.\ \textAR{والدة}), counts must be inferred from number-noun constructions (\textAR{ابنين} ``two sons''), and relationship types must be disambiguated (\textAR{أخ} can denote full, paternal, or maternal brother). This entity extraction and relationship parsing from unstructured text represents a core NLP challenge absent in symbolic approaches.

\item \textbf{Conditional logic through learned patterns}: Share assignments depend on presence/absence of other heirs (e.g., wife receives 1/4 if no children exist, 1/8 otherwise). Unlike rule-based systems where such conditions are explicitly programmed, neural models must learn these conditional dependencies from training examples. With only finite data and class imbalances (e.g., \textit{radd} appears in 2.8\% of cases), models must generalize learned patterns to unseen combinations of heir configurations.

\item \textbf{Hierarchical blocking through pattern recognition}: Distant relatives are excluded by closer ones following precedence rules (e.g., sons block grandsons, fathers block uncles). Models must learn these hierarchical relationships from examples rather than executing explicit genealogical graphs, requiring pattern recognition over complex family structures with multiple generations.

\item \textbf{Conditional algorithm selection}: The model must detect which distribution algorithm to apply based on computed share totals. Standard cases (92.3\%) use residuary distribution when a male agnate heir (\textAR{عصبة}) exists, \textit{'awl} cases (4.9\%) require proportional reduction when shares exceed unity, and \textit{radd} cases (2.8\%) require surplus redistribution when shares sum to less than unity and no residuary heir is present (see section~\ref{sec:dataset}). Correct detection requires both arithmetic computation (checking if shares sum to $<$1, $=$1, or $>$1) and logical reasoning (verifying absence of residuary heirs). The statistical rarity of adjustment cases in training data exacerbates the learning challenge.

\item \textbf{Numerical precision in text generation}: Unlike symbolic systems that perform exact fractional arithmetic using rational number representations, neural models must generate fractions and percentages as text strings while maintaining numerical correctness. All percentages must sum to exactly 100\%, fractions must be in lowest terms, and floating-point approximations are invalid. This requires learning precise numerical patterns from examples rather than executing deterministic calculations.

\item \textbf{Structured output generation}: Models must produce syntactically valid JSON with consistent Arabic terminology, proper nesting, and exact schema compliance. Errors in JSON syntax (missing brackets, unclosed quotes), inconsistent heir naming across sections, or schema violations invalidate the entire output. This generation constraint requires maintaining structural coherence across potentially long output sequences.
\end{itemize}

The task difficulty arises not from the logical complexity of inheritance rules themselves---which are well-defined and deterministic---but from the requirement to learn and apply these rules through neural pattern matching on natural language inputs while generating structured outputs with exact numerical precision. Errors at any intermediate stage propagate deterministically and invalidate the final distribution. A correct solution requires models to simultaneously excel at linguistic understanding, knowledge-intensive reasoning, numerical computation, and constrained generation without access to gold intermediate representations or symbolic verification mechanisms. This makes \textit{MAWARITH} a demanding testbed for evaluating whether neural models can approximate rule-based reasoning through end-to-end learning from examples.

\subsection{MAWARITH Dataset}
\label{sec:dataset}
The \textit{MAWARITH} dataset~\cite{bouchekif2025mawarith} comprises 12,500 Arabic inheritance cases following the majority opinion (\textAR{الجمهور}, \textit{al-jumh\=ur}). The corpus is split into 12,000 training instances and 500 test instances, covering 36 distinct heir categories ranging from close relatives (parents, children, spouses) to distant extended family across multiple generations.

Table~\ref{tab:dataset_stats} shows the dataset composition. The majority (92.3\%) are simple cases requiring no adjustment, while 4.9\% involve \textAR{عول} (\textit{'awl}, proportional reduction) and 2.8\% involve \textAR{رد} (\textit{radd}, surplus redistribution).

\begin{table}[h]
\centering
\begin{tabular}{lrrrr}
\toprule
\textbf{Split} & \textbf{Simple} & \textbf{'awl} & \textbf{radd} & \textbf{Total} \\
\midrule
Training & 11,079 & 577 & 344 & 12,000 \\
Test     &    456 &  39 &   5 &    500 \\
\midrule
Total    & 11,535 & 616 & 349 & 12,500 \\
\bottomrule
\end{tabular}
\caption{MAWARITH dataset structure with distribution of inheritance cases by complexity.}
\label{tab:dataset_stats}
\end{table}

\begin{table*}[t]
\centering
\small
\begin{tabularx}{\textwidth}{lX}
\toprule
\textbf{Field} & \textbf{Content} \\
\midrule

Question & \textAR{مات وترك: عم لأب و ابن أخ لأب و أربع بنات ابن و أم الأم و أب الأب و زوجة و خمسة أبناء ابن أخ لأب. ما هو نصيب كل وريث؟} \\
 & (A person died leaving: a paternal uncle, a son of a paternal brother, four sons' daughters, the maternal grandmother, the paternal grandfather, a wife, and five sons of a paternal nephew. What is each heir's share?) \\


Heirs & \texttt{[}
\textAR{وريث: أم الأم، عدد: 1}،
\textAR{وريث: زوجة، عدد: 1}،
\textAR{وريث: أب الأب، عدد: 1}،
\textAR{وريث: بنت ابن، عدد: 4}
\texttt{]} \\
 & (heir: maternal grandmother, count: 1; heir: wife, count: 1; heir: paternal grandfather, count: 1; heir: son’s daughter, count: 4) \\


Blocked & \texttt{[}
\textAR{وريث: عم لأب، عدد: 1}،
\textAR{وريث: ابن أخ لأب، عدد: 1}،
\textAR{وريث: ابن ابن أخ لأب، عدد: 5}
\texttt{]} \\
 & (heir: paternal uncle, count: 1; heir: son of a paternal brother, count: 1; heir: sons of a son of a paternal brother, count: 5) \\


Shares & \texttt{[}
\textAR{وريث: أم الأم، عدد: 1، كسر: 1/6}،
\textAR{وريث: زوجة، عدد: 1، كسر: 1/8}،
\textAR{وريث: أب الأب، عدد: 1، كسر: 1/6}،
\textAR{وريث: بنت ابن، عدد: 4، كسر: 2/3}
\texttt{]} \\
 & (heir: maternal grandmother, count: 1, fraction: 1/6; heir: wife, count: 1, fraction: 1/8; heir: paternal grandfather, count: 1, fraction: 1/6; heir: son’s daughters, count: 4, fraction: 2/3) \\


\textit{'Awl} or \textit{radd} &
\textAR{عول} \\
& (\textit{'Awl} --- proportional reduction, as shares sum to
$\dfrac{1}{6}+\dfrac{1}{8}+\dfrac{1}{6}+\dfrac{2}{3}
= \dfrac{27}{24} > 1$) \\


Post-Tasil & \textAR{مجموع الأسهم: 72، توزيع: زوجة 3/72، أب الأب 4/72، بنت ابن 4/72 لكل واحدة، أم الأم 4/72} \\
 & (total shares: 27; distribution: wife 3/27, paternal grandfather 4/27, each son’s daughter 4/27, maternal grandmother 4/27) \\

\bottomrule
\end{tabularx}

\caption{Example inheritance case from MAWARITH with Arabic text and full English translation. The prescribed shares exceed the estate, so the case undergoes \textAR{عول} (proportional reduction). The four \textAR{بنات ابن} \textit{(son’s daughter)} share 16/27 collectively, giving each one 4/27.}

\label{tab:example}
\end{table*}

Table~\ref{tab:example} shows a simplified example from the training data. This case demonstrates \textit{'awl} (proportional reduction) where prescribed shares exceed the estate. The gold standard provides a human-readable explanation following Islamic legal reasoning traditions, while the structured \texttt{JSON} output enables automated evaluation via the MIR-E metric, with each field corresponding to a distinct reasoning stage assessed independently.

\subsection{Domain Adaptation Data}

For stage 1 domain adaptation, we use 3,166 Islamic fatwa records from Islamweb\footnote{\url{https://www.islamweb.net/}} covering inheritance-related religious rulings. These fatwas provide natural-language explanations of inheritance scenarios, introducing models to jurisprudential terminology (\textAR{الورثة} \textit{al-waratha}, \textAR{الحجب} \textit{al-\d{h}ajb}, \textAR{العصبة} \textit{al-'a\d{s}aba}) and reasoning patterns used by Islamic legal scholars.

\section{Methodology}

\subsection{Model Architecture}

We use Qwen3-4B~\cite{Qwen2025} as our base model. Qwen3 is a multilingual reasoning model trained on diverse corpora including Arabic text, achieving strong performance on mathematical and logical reasoning benchmarks while maintaining a compact parameter count suitable for resource-constrained settings.

\subsection{Multi-Stage QLoRA Fine-Tuning}

Our training approach consists of two stages. Table~\ref{tab:training_config} explains the training configuration for the two phases of the proposed multi-stage QLoRA fine-tuning approach.

\begin{table*}[t]
\centering
\small
\begin{tabular}{lcc}
\toprule
\textbf{Configuration} & \textbf{Phase 1: Domain Adaptation} & \textbf{Phase 2: Task-Specific Training} \\
\midrule
Training data & 3,166 Islamic fatwa records & 12,000 inheritance cases \\
Objective & Causal language modeling & Structured JSON generation \\
Learning rate & $2 \times 10^{-4}$ & $3 \times 10^{-5}$ \\
Epochs & 2 & 6 \\
Batch size & 1 & 1 \\
Gradient accumulation & 8 steps & 8 steps \\
Max sequence length & 2048 & 2048 \\
Optimizer & AdamW & AdamW \\
Warmup ratio & 0.1 & 0.1 \\
Base model & Qwen3-4B & Qwen3-4B \\
Quantization & 4-bit NF4 & 4-bit NF4 \\
LoRA rank ($r$) & 128 & 128 \\
LoRA alpha ($\alpha$) & 256 & 256 \\
Compute dtype & bfloat16 & bfloat16 \\
Target modules & All projection layers & All projection layers \\
\bottomrule
\end{tabular}
\caption{Training configuration for the two phases of the proposed multi-stage QLoRA fine-tuning approach. Phase 1 performs domain adaptation on Islamic fatwa records, while Phase 2 focuses on structured inheritance reasoning and JSON output generation.}
\label{tab:training_config}
\end{table*}

\subsubsection{Stage 1: Domain Adaptation}

In the first stage, we fine-tune Qwen3-4B on 3,166 Islamic fatwa records to acquire inheritance-specific terminology and jurisprudential reasoning patterns. This stage uses a standard causal language modeling objective where the model learns to generate fatwa explanations given inheritance questions.

\subsubsection{Stage 2: Task-Specific Training}

The second stage continues training on 12,000 structured inheritance cases, teaching the model to produce JSON-formatted outputs with correct heir identification, blocking decisions, share calculations, and adjustment mechanisms.

We filter training examples to include only those with valid JSON outputs, yielding 12,000 usable instances. The system prompt explicitly instructs the model to output JSON only without additional explanation:

\begin{quote}
\small
\noindent``Output \texttt{JSON} only without any additional text. Do not write
explanation, thinking, or symbols outside \texttt{JSON}. The \texttt{JSON}
must contain only the following keys: \texttt{heirs, blocked, shares,
awl\_or\_radd, post\_tasil}.''
\end{quote}

\subsection{QLoRA Configuration}

Both training stages employ QLoRA~\cite{dettmers2023qlora} for parameter-efficient fine-tuning. We adopt 4-bit NF4 quantization with double quantization to reduce memory usage while maintaining performance. The LoRA configuration uses a rank of $r = 128$ and a scaling factor of $\alpha = 256$. Adaptation is applied to all projection layers, including \texttt{q\_proj}, \texttt{k\_proj}, \texttt{v\_proj}, \texttt{o\_proj}, \texttt{gate\_proj}, \texttt{up\_proj}, and \texttt{down\_proj}. All computations are performed using the bfloat16 data type to balance numerical stability and efficiency.

\section{Evaluation Metric}

\subsection{MIR-E: Mawarith Inheritance Reasoning Evaluation}

The QIAS 2026 shared task proposes MIR-E (Mawarith Inheritance Reasoning Evaluation)~\cite{bouchekif2025mawarith}, a weighted multi-stage evaluation metric designed to assess both intermediate reasoning steps and final outputs in Islamic inheritance problems. Unlike traditional evaluation methods that focus only on final answers, MIR-E enables fine-grained assessment of structured reasoning by decomposing the inheritance process into key stages.

MIR-E evaluates model predictions across four components:

\begin{enumerate}
\item \textbf{Heirs and Blocking ($S_h$)}: This component evaluates whether the model correctly identifies the set of effective heirs after applying blocking rules. It combines an F1 score over the predicted and gold effective heir sets with count accuracy, penalizing missing heirs, spurious heirs, incorrect blocking decisions, and count mismatches.

\item \textbf{Share Assignment ($S_s$)}: This component measures the correctness of the assigned shares for all eligible heirs. Predicted shares are compared against gold values using a small tolerance threshold to account for minor numerical deviations.

\item \textbf{Adjustment ($S_a$)}: This component evaluates whether the model correctly identifies the required adjustment type (none, \textAR{عول}, or \textAR{رد}). Since adjustment depends on earlier reasoning stages, it is evaluated conditionally and assigned a non-zero score only when both $S_h = 1$ and $S_s = 1$.

\item \textbf{Final Allocation ($S_f$)}: This component measures the accuracy of the final normalized distribution of shares across heirs after completing the full inheritance computation. Predictions are evaluated using a tolerance threshold.
\end{enumerate}

The overall MIR-E score is computed as a weighted sum of these components:

\begin{equation}
\text{MIR-E} = \alpha_h S_h + \alpha_s S_s + \alpha_a S_a + \alpha_f S_f
\end{equation}

where $\alpha_h = \alpha_s = \alpha_f = 0.30$ and $\alpha_a = 0.10$. Equal weights are assigned to heir identification, share assignment, and final allocation, as these stages capture the core reasoning process, while the adjustment component receives a lower weight due to its conditional nature and lower frequency.

This evaluation framework enables detailed error analysis by isolating failures at different stages of the inheritance reasoning pipeline, including heir identification, share computation, adjustment detection, and final allocation.

\begin{table*}[h]
\centering
\small
\setlength{\tabcolsep}{6pt}
\begin{tabular}{lcccc}
\toprule
\textbf{Model} & $S_h$ & $S_s$ & $S_a$ & \textbf{MIR-E} \\
\midrule
Gemini-2.5-flash       & 94.5\% & 92.9\% & 89.4\% & \textbf{90.1\%} \\
\textbf{QU-NLP (Ours)} & \textbf{97.1\%} & \textbf{94.3\%} & 84.6\% & \textbf{90.0\%} \\
\midrule
Qwen3-32B              & 69.0\% & 44.6\% & 26.5\% & 43.7\% \\
GPT-OSS-120B           & 69.3\% & 32.7\% & 27.1\% & 39.1\% \\
LLaMA-3.3-70B          & 64.8\% & 40.3\% & 21.5\% & 39.0\% \\
Fanar-Sadiq            & 62.1\% & 36.7\% & 20.4\% & 36.8\% \\
Fanar-C-2-27B          & 58.4\% & 31.4\% & 17.8\% & 32.8\% \\
\bottomrule
\end{tabular}
\caption{Component-wise MIR-E scores on the 500-case test set after
post-processing. $S_h$: heirs and blocking, $S_s$: share assignment,
$S_a$: adjustment detection.
QU-NLP results reflect the Basic post-processing pipeline.}
\label{tab:results}
\end{table*}

\section{Experimental Setup}

\subsection{Implementation Details}

We implement our approach using the Hugging Face Transformers library~\cite{wolf2020transformers} and the PEFT library for parameter-efficient fine-tuning~\cite{peft}. Inference is performed using a Qwen3-4B base model with a LoRA adapter, with automatic device placement.

\textbf{Data Preprocessing:} Inputs are formatted using the Qwen3 chat template with system and user roles. A fixed system prompt enforces strict JSON-only outputs with predefined keys corresponding to the inheritance reasoning schema.

\textbf{Generation Parameters:} During inference, we use greedy decoding with temperature set to 0.0, a maximum of 1024 generated tokens, and no sampling.

\textbf{Post-Processing:} Model outputs are post-processed using a multi-stage pipeline. First, raw outputs are cleaned by removing \texttt{<think>} tags and extracting the JSON structure. Additional normalization includes typo correction, removal of Arabic elongation characters, and structural validation. We further apply rule-based corrections, including deduplication of blocked heirs, normalization of \texttt{awl\_or\_radd} labels, and recalculation of post-\textit{ta\=s\={\i}l} distributions using exact fraction arithmetic when necessary. The pipeline implements a neural-symbolic separation: the neural component handles jurisprudential reasoning while the symbolic component verifies adjustment arithmetic using exact rational arithmetic, as evidenced by the PostTasil variant producing results identical to Basic across all metrics (Section~\ref{sec:postproc}).

\subsection{Baseline Comparisons}

We compare our approach against a set of six large language models evaluated in prior work~\cite{bouchekif2025mawarith}, spanning Arabic-specialized systems, open-weight multilingual models, and commercial reasoning models. These include the Arabic-centric \textit{Fanar} models, evaluated in both their general-purpose variant \textit{Fanar-C-2-27B} and Islamic-specialized variant \textit{Fanar-Sadiq}, accessed via the Fanar API\footnote{\url{https://api.fanar.qa/docs\#description/introduction}}. We also include open-weight multilingual models, namely \textit{LLaMA-3.3-70B}\footnote{\url{https://huggingface.co/meta-llama/Meta-Llama-3-70B}} and \textit{GPT-OSS-120B}\footnote{\url{https://huggingface.co/openai/gpt-oss-120b}}, both accessed through the Groq API\footnote{\url{https://console.groq.com/}}. Additionally, we evaluate \textit{Qwen3-32B}\footnote{\url{https://huggingface.co/Qwen/Qwen3-32B}}, a multilingual reasoning model. Finally, we include \textit{Gemini-2.5-flash}, a commercial reasoning model.

All baseline systems are evaluated in a zero-shot setting using a unified Arabic prompt that specifies the inheritance scenario, required reasoning steps, and the target JSON output schema, ensuring a fair comparison across models without task-specific fine-tuning. In contrast, our model uses a task-specific inference prompt implemented through the Qwen chat template, consisting of a fixed system instruction that enforces strict JSON-only outputs with predefined keys, followed by the user query.

\section{Results and Discussion}

\subsection{Overall Performance}

Table~\ref{tab:results} presents overall MIR-E scores and component-wise performance on the 500-instance test set. Our QU-NLP system achieves 90.0\% MIR-E, closely matching the commercial Gemini-2.5-flash model (90.1\%) and substantially outperforming all open-weight baselines. Notably, our 4B parameter model outperforms systems with 8--30$\times$ more parameters (Qwen3-32B, LLaMA-3.3-70B, GPT-OSS-120B), demonstrating that domain-specific fine-tuning on high-quality structured data is more effective than relying solely on model scale and general reasoning capabilities.

\subsection{Effect of Post-Processing}
\label{sec:postproc}

Beyond model training, we apply a lightweight post-processing pipeline to raw predictions before evaluation. The pipeline operates in three stages: (1) typographic normalisation, correcting Arabic elongation characters (tatweel) and common spelling variants such as \textAR{باقى}~$\rightarrow$~\textAR{باقي}; (2) structural deduplication, removing any heir that appears simultaneously in both \texttt{heirs} and \texttt{blocked}, since Islamic jurisprudence forbids an individual from occupying both roles; and (3) label normalisation, replacing unrecognised \texttt{awl\_or\_radd} strings with a value inferred from the fraction sum, while leaving valid labels (\textAR{عول}, \textAR{رد}, \textAR{لا}) unchanged. A fourth variant, \emph{PostTasil}, additionally attempts to recalculate the final distribution table when the model's \texttt{post\_tasil} fractions are identical to the unadjusted \texttt{shares}, indicating the model omitted the \textAR{عول}/\textAR{رد} adjustment step.

Table~\ref{tab:postprocessing} reports the impact of each variant on all 500 evaluated test cases.

\begin{table}[h]
\centering
\small
\setlength{\tabcolsep}{5pt}
\begin{tabular}{lccc}
\toprule
\textbf{Metric} & \textbf{Original} & \textbf{Basic} & \textbf{PostTasil} \\
\midrule
MIR-E Overall        & 89.8\% & \textbf{90.0\%} & \textbf{90.0\%} \\
Heirs \& Blocking    & 96.4\% & \textbf{97.1\%} & \textbf{97.1\%} \\
Share Assignment     & 94.3\% & 94.3\%          & 94.3\%          \\
'Awl/Radd Detection  & 84.6\% & 84.6\%          & 84.6\%          \\
Final Distribution   & 80.5\% & 80.5\%          & 80.5\%          \\
\bottomrule
\end{tabular}
\caption{Effect of post-processing on component scores across 500 test cases. \emph{Original}: raw model output. \emph{Basic}: typographic cleaning and structural deduplication. \emph{PostTasil}: Basic plus recalculation of final distributions for \textAR{عول}/\textAR{رد} cases. Bold denotes the highest score per metric.}
\label{tab:postprocessing}
\end{table}

Post-processing yields a net gain of $+0.2$pp on overall MIR-E (89.8\%~$\rightarrow$~90.0\%), with the improvement concentrated entirely in the Heirs~\&~Blocking component ($+0.7$pp). This gain is attributable to the deduplication step: across the test set, a subset of predictions placed the same individual in both \texttt{heirs} and \texttt{blocked}---a structural contradiction that the Basic pipeline resolves by trusting the \texttt{heirs} list.

The Share Assignment, 'Awl/Radd Detection, and Final Distribution components are unaffected by post-processing, which is intentional. Share fractions are taken directly from the model's output without arithmetic intervention, as any correction would require reconstructing the underlying jurisprudential reasoning. The \texttt{awl\_or\_radd} label is normalised only for unrecognised strings; valid labels are always preserved. Overriding valid labels based on fraction sums risks conflating two distinct error types: a model that correctly identifies \textAR{لا} \textit{(no)} but assigns a wrong fraction to a residuary heir will produce an inflated sum, yet its \texttt{awl\_or\_radd} label is correct. This motivated the conservative design decision to trust all valid labels unconditionally, applying correction only when the model produces an unrecognised label.

The PostTasil variant produces results identical to Basic across all metrics, indicating that in cases where the model applies \textAR{عول} (\textit{'awl}) or \textAR{رد} (\textit{radd}), its \texttt{post\_tasil} distribution already differs from the unadjusted \texttt{shares}---the recalculator's trust condition is not met, so the model's own output is preserved. This confirms the neural-symbolic separation in our pipeline: the symbolic verifier found no cases requiring arithmetic correction, demonstrating that the model's adjustment arithmetic is correct in every case it correctly classifies the adjustment type. The 29 calculation errors (5.8\%) are therefore confined to the final percentage generation step---cases where the model reasons correctly through all five stages but generates numerically inconsistent text in the output field (see section~\ref{sec:error}).

\subsection{Component-Wise Analysis}

\textbf{Heir Identification ($S_h = 97.1\%$):} Our model achieves the highest heir identification score among all evaluated systems, exceeding Gemini-2.5-flash by 2.6pp (97.1\% vs.\ 94.5\%) and surpassing all open-weight baselines by a wide margin (58.4\%--69.3\%).

\textbf{Share Assignment ($S_s = 94.3\%$):} Share calculation accuracy exceeds Gemini-2.5-flash by 1.4pp (94.3\% vs.\ 92.9\%), and substantially surpasses all open-weight baselines (31.4\%--44.6\%). The multi-stage training strategy allows the model to first acquire share terminology and fractional notation in Stage~1, then practise accurate assignment under varied heir configurations in Stage~2.

\textbf{Adjustment Detection ($S_a = 84.6\%$):} Detecting the required adjustment type is inherently challenging: \textAR{عول} (\textit{'awl}, proportional reduction) requires correctly summing all assigned fractions and identifying when they exceed unity, while \textAR{رد} (\textit{radd}, surplus redistribution) additionally requires confirming that no residuary heir is present to absorb the surplus---a condition the model must infer from the absence of certain heir types rather than their presence. The rarity of both adjustment types in training, \textAR{عول} at 4.9\% and \textAR{رد} at 2.8\% of cases, limits exposure to these scenarios and contributes to systematic underfitting. Consequently, our model scores 84.6\%, trailing Gemini-2.5-flash by 4.8pp (89.4\%) but substantially outperforming all open-weight baselines, which range from 17.8\% to 27.1\%.

Per-category analysis across the 500 evaluated predictions further illuminates adjustment performance. Table~\ref{tab:per_category} shows MIR-E broken down by case type. Counter-intuitively, \textit{'awl} cases perform worse than \textit{radd} cases (79.2\% vs.\ 83.0\%) despite having nearly eight times more training examples (577 vs.\ 344 training, 39 vs.\ 5 test). This reversal reveals that performance is limited by arithmetic complexity rather than data frequency: \textit{'awl} requires computing a new common denominator from multiple overlapping fractions ($S_a = 66.7\%$, $S_f = 50.4\%$), while \textit{radd} requires only proportional redistribution once the type is identified. For \textit{radd} specifically, heir identification and share assignment both reach 100\%---legal reasoning is perfect on all five test cases, with failures confined to one detection error and two final distribution arithmetic failures. The perfect $S_h$ and $S_s$ scores on \textit{radd} cases argue directly against the memorisation hypothesis: a model relying on pattern matching over 344 training instances would be expected to fail on legal reasoning first, not exclusively on the arithmetic output step.

\begin{table*}[h]
\centering
\small
\begin{tabular}{lrrrrrr}
\toprule
\textbf{Type} & \textbf{N} & \textbf{MIR-E} & $S_h$ & $S_s$ & $S_a$ & $S_f$ \\
\midrule
Simple & 456 & 90.8\% & 96.4\% & 94.1\% & 86.2\% & 83.5\% \\
\textAR{عول} (\textit{'awl}) & 39 & 79.2\% & 96.2\% & 95.1\% & 66.7\% & 50.4\% \\
\textAR{رد} (\textit{radd}) &  5 & 83.0\% & 100.0\% & 100.0\% & 80.0\% & 50.0\% \\
\bottomrule
\end{tabular}
\caption{Per-category MIR-E scores. \textit{'Awl} cases underperform \textit{radd} cases (79.2\% vs.\ 83.0\%) despite nearly eight times more training examples, revealing arithmetic complexity rather than data frequency as the primary bottleneck.}
\label{tab:per_category}
\end{table*}

\subsection{Error Analysis}
\label{sec:error}
To understand our model's limitations, we conduct detailed error analysis on all 500 test cases, comparing model predictions against gold standard references across all reasoning components. We identify four distinct error categories that account for 16\% of test cases.

\subsubsection{Error Distribution}

Table~\ref{tab:error_dist} shows the distribution of errors by category. Calculation errors represent the most impactful failure mode ($-1.7$pp on MIR-E), followed by residue label avoidance ($-0.85$pp), heir identification ($-0.4$pp), and radd detection ($-0.2$pp).

\begin{table}[h]
\centering
\small
\begin{tabular}{lp{0.6cm}p{0.6cm}p{1cm}}
\toprule
\textbf{Error Type} & \textbf{Cases} & \textbf{Total} & \textbf{Impact} \\
\midrule
Calculation                    &  29 &  5.8\% & $-$1.7pp \\
Residue label avoidance$^\dagger$ & 314 & 62.9\% & $-$0.85pp \\
Heir Identification            &  19 &  3.8\% & $-$0.4pp \\
Radd Detection                 &  11 &  2.2\% & $-$0.2pp \\
\bottomrule
\end{tabular}
\caption{Distribution of error types with impact on overall MIR-E.
Calculation, heir identification, and radd detection are discrete failure
cases (59 total, 11.8\% of test set); impacts are estimated.
$^\dagger$Among the 417 cases where the gold solution requires
\textAR{باقى التركة}, the model substitutes an explicit fraction in 314
(75.3\%). Because 83.1\% of those 314 cases write the numerically correct
fraction, MIR-E's tolerance absorbs most of the penalty: the measured
shares score gap between cases with and without the label is 0.045,
giving a global cost of 314/500 x 0.045 x 0.30 = $-0.85$pp (shares score: 0.97 with label vs.\ 0.92 without, gap = 0.045).
This row is not summed with the three discrete error categories above.}
\label{tab:error_dist}
\end{table}
 
\subsubsection{Representative Error Cases}

Table~\ref{tab:error_cases} presents component scores for one
representative case from each error category.

\begin{table*}[h]
\centering
\small
\begin{tabular}{llccccc}
\toprule
\textbf{Case ID} & \textbf{Error Type} & $S_h$ & $S_s$ & $S_a$ & $S_f$ & \textbf{MIR-E} \\
\midrule
nf8w4p3x\_7  & Calculation      & 1.00 & 1.00 & 1.00 & 0.00 & 0.700 \\
nf5n5k1z\_7  & Heir ID          & 0.66 & 1.00 & 1.00 & 1.00 & 0.898 \\
ng7z4j2b\_2  & Radd Detection   & 1.00 & 1.00 & 0.00 & 0.00 & 0.600 \\
ng2p5t2e\_10 & Share Assignment & 0.81 & 0.67 & 0.00 & 0.33 & 0.542 \\
\bottomrule
\end{tabular}
\caption{Component scores for representative error cases
($S_h$: heirs/blocking, $S_s$: shares, $S_a$: adjustment,
$S_f$: final distribution).}
\label{tab:error_cases}
\end{table*}

\textbf{Case 1 -- Calculation Error (nf8w4p3x\_7):}
A deceased left a paternal half-uncle (\textAR{عم لأب}), a son of a paternal half-brother (\textAR{ابن أخ لأب}), four sons' daughters (\textAR{بنت ابن}), a maternal grandmother (\textAR{أم الأم}), a paternal grandfather (\textAR{أب الأب}), a wife (\textAR{زوجة}), and five sons of a paternal nephew (\textAR{ابن ابن أخ لأب}). The model correctly identifies the four eligible heirs and blocks the remaining three, assigns correct shares ($\frac{1}{6}, \frac{1}{8}, \frac{1}{6}, \frac{2}{3}$), and correctly detects \textAR{عول} (\textit{'awl}) since shares sum to $\frac{27}{24} > 1$, giving awl denominator~27 ($S_h = S_s = S_a = 1$). However, \texttt{post\_tasil} uses \texttt{total\_shares = 108} ($= 27 \times 4$, conflating the awl denominator with the count of \textAR{بنت ابن}) and copies pre-awl values, producing wife = 8.33\% instead of 11.11\% ($-2.78$pp) and all other heirs at 13.89\% instead of 14.81\% ($-0.93$pp), summing to only 91.67\%. Legal reasoning is perfect; the error is a failure to propagate the \textit{'awl} adjustment through to the final output field.

\textbf{Case 2 -- Heir Identification Error (nf5n5k1z\_7):}
A deceased left a paternal great-grandfather (\textAR{أب أب الأب}), two paternal half-brothers (\textAR{أخ لأب}), a maternal grandmother (\textAR{أم الأم}), a father (\textAR{أب}), two of the father's paternal half-uncles (\textAR{عم الأب لأب}), and two great-grandmothers (\textAR{أم أم الأب}, \textAR{أم أب الأب}). The gold solution correctly identifies two heirs (\textAR{أم الأم} with $\frac{1}{6}$ and \textAR{أب} with the residue) and blocks the remaining five. The model's heirs, shares, and final distribution are all correct ($S_s = S_f = 1$), and \texttt{post\_tasil} matches gold exactly (16.67\% and 83.33\%). The $S_h$ penalty (0.66) arises solely from a label mismatch in the blocked list: the model writes \textAR{عم الأب} instead of the gold taxonomy label \textAR{عم الأب لأب} for the father's paternal half-uncle. The underlying blocking decision is correct, but the shortened label fails the string-matching evaluation.

\textbf{Case 3 -- Radd Detection Error (ng7z4j2b\_2):}
A deceased left three full sisters (\textAR{أخت شقيقة}) and a paternal great-grandmother (\textAR{أم أب الأب}). The model correctly identifies both heirs and assigns correct shares (\textAR{أم أب الأب}: $\frac{1}{6}$, \textAR{أخت شقيقة}: $\frac{2}{3}$). Since the shares sum to $\frac{5}{6} < 1$ and no residuary heir is present, \textAR{رد} (\textit{radd}) should apply. The model predicts \textAR{لا} (no adjustment), failing to confirm the absence of a residuary heir~\cite{bouchekif2025mawarith}. Without redistribution the final percentages are wrong: \textAR{أم أب الأب} receives 16.67\% instead of 20.00\% ($+3.33$pp) and each \textAR{أخت شقيقة} receives 22.22\% instead of 26.67\% ($+4.44$pp). Since \textAR{رد} cases are only 2.8\% of training data, this reflects systematic underfitting on rare adjustment events.

\textbf{Case 4 -- Share Assignment Error (ng2p5t2e\_10):}
A deceased left a paternal grandfather (\textAR{أب الأب}), four sons of a full paternal uncle (\textAR{ابن عم شقيق}), five sons of a paternal cousin (\textAR{ابن ابن عم شقيق}), four paternal half-brothers (\textAR{أخ لأب}), a mother (\textAR{أم}), a maternal half-brother (\textAR{أخ لأم}), a paternal grandmother (\textAR{أم الأب}), a son of a paternal half-uncle (\textAR{ابن عم لأب}), a maternal grandmother (\textAR{أم الأم}), and four of the father's uncles (\textAR{عم الأب}). The gold solution has three heirs: \textAR{أم} ($\frac{1}{6}$), \textAR{أب الأب} ($\frac{5}{18}$), and \textAR{أخ لأب}$\times 4$ (residue, \textAR{باقى التركة}). The model makes three errors. First, it incorrectly adds \textAR{أخ لأم} (\textit{maternal half-brother}) as an eligible heir, who is blocked when the mother is present. Second, it lists four full brothers (\textAR{أخ شقيق}) as blocked---a relative not mentioned anywhere in the question, a hallucinated heir type. Third, it assigns \textAR{أب الأب} a share of $\frac{1}{3}$ instead of $\frac{5}{18}$ and replaces the residue designation for \textAR{أخ لأب} with an explicit fraction of $\frac{1}{6}$---a systematic avoidance of \textAR{باقى التركة}. The result: \textAR{أب الأب} receives 16.67\% instead of 27.78\% ($-11.11$pp), each \textAR{أخ لأب} receives 8.33\% instead of 13.89\% ($-5.56$pp), and the spurious \textAR{أخ لأم} absorbs an additional 16.67\% of the estate, with the prediction summing to only 83.33\% of the estate.

\subsubsection{Error Patterns and Implications}

\textbf{Arithmetic vs. Semantic Errors:} The most impactful errors (calculation, 29 cases, see Table~\ref{tab:error_dist}) are \textit{non-semantic}---the model understands inheritance law correctly but fails in final arithmetic. This suggests errors occur in the text-generation step of the final output field rather than in core reasoning, making them addressable through constrained decoding without retraining.

\textbf{Complexity Degradation:} Performance varies with case complexity. For simple cases involving 2--4 heirs, the model achieves approximately 91.7\% MIR-E. For medium-complexity cases with 5--7 heirs, performance decreases slightly to around 88.9\%, while for complex cases involving $\geq$8 mentioned heirs, it further declines to approximately 87.0\%. This gradual degradation indicates a moderate impact of complexity on performance, with the model maintaining strong accuracy even in more complex scenarios.

\textbf{Rare Event Underfitting:} Per-category analysis (Table~\ref{tab:per_category}) shows that \textit{'awl} cases underperform \textit{radd} cases (79.2\% vs.\ 83.0\%) despite having nearly eight times more training examples. The primary bottleneck is arithmetic complexity: \textit{'awl} requires multi-fraction common-denominator computation, while \textit{radd} requires only proportional redistribution once the type is identified. Both benefit from oversampling and explicit rule-based fallbacks as future work.

\textbf{Residue Label Recall:} The gold standard requires the residuary label \textAR{باقى التركة} in 417 of 500 evaluated cases (83.4\%), reflecting the prevalence of male agnate heirs (\textAR{عصبة}) across the test set. The model provides this label in only 103 of those cases (24.7\% recall), substituting an explicit fraction in the remaining 314 cases (75.3\% avoidance rate). Table~\ref{tab:residue} summarises the breakdown.

\begin{table*}[h]
\centering
\small
\begin{tabular}{lrr}
\toprule
\textbf{Residue label behaviour} & \textbf{Cases} & \textbf{Rate} \\
\midrule
Gold requires residue label      & 417 & 83.4\% of test \\
Model provides label (recall)    & 103 & 24.7\% of required \\
Model avoids label               & 314 & 75.3\% of required \\
\quad of which: correct fraction & 261 & 83.1\% of avoided \\
\quad of which: wrong fraction   &  53 & 16.9\% of avoided \\
\midrule
Global MIR-E cost & \multicolumn{2}{r}{$-0.85$pp} \\
\bottomrule
\end{tabular}
\caption{Residue label recall analysis. Despite a 75.3\% avoidance rate, the global MIR-E cost is only $-0.85$pp because 83.1\% of avoidance cases compute the numerically correct fraction within the evaluation tolerance.}
\label{tab:residue}
\end{table*}

Despite the low recall, the MIR-E cost is only $-0.85$pp because MIR-E's tolerance absorbs 83.1\% of avoidance cases: the model computes the numerically correct residue fraction and writes it as an explicit value---for instance, writing \texttt{"7/12"} when \textAR{باقى التركة} is expected, which falls within the evaluation tolerance. This reveals a \textit{representational} rather than \textit{computational} failure. Among the 314 avoidance cases, 261 (83.1\%) produce the exact fraction a symbolic calculator would derive by subtracting fixed shares from unity---the model has learned to perform residue arithmetic correctly but defaults to explicit fraction notation due to training bias toward fixed-share cases. The failure is therefore in the final token selection, not in the underlying calculation. Constrained decoding enforcing \textAR{باقى التركة} whenever an \textAR{عصبة} heir is present and fixed shares sum to less than unity would recover the correct label in the majority of affected cases without any change to the model weights.

\subsection{Pipeline Success Rate}

Table~\ref{tab:pipeline} reports cumulative success rates across the
reasoning pipeline, where each row shows the percentage of cases in
which all stages up to and including that point score perfectly.

\begin{table}[h]
\centering
\small
\begin{tabular}{p{4.8cm}p{2cm}}
\toprule
\textbf{Reasoning Stage} & \textbf{Success Rate} \\
\midrule
$S_h = 1$ (Heirs correct)                     & 84.0\% \\
$S_h = 1,\ S_s = 1$ (+ Shares correct)        & 81.2\% \\
$S_h = 1,\ S_s = 1,\ S_a = 1$ (+ Adjustment) & 79.4\% \\
All stages correct                             & 65.5\% \\
\bottomrule
\end{tabular}
\caption{Cumulative pipeline success rates. Each row shows the
percentage of cases where all stages up to and including that
point score perfectly ($S = 1$).}
\label{tab:pipeline}
\end{table}

While 65.5\% of cases are solved perfectly across all components, the overall MIR-E reaches 90.0\%. This gap is explained by the partial-credit design of MIR-E~\cite{bouchekif2025mawarith}, which assigns weighted scores to each intermediate stage ($\alpha_h = \alpha_s = \alpha_f = 0.30$, $\alpha_a = 0.10$). A case with correct heirs, shares, and adjustment but wrong final percentages, for instance, still earns $0.30 + 0.30 + 0.10 + 0.00 = 0.70$ MIR-E. It is worth noting that the adjustment score $S_a$ is evaluated conditionally: it receives a non-zero value only when both $S_h = 1$ and $S_s = 1$, reflecting the sequential dependency of the reasoning pipeline. The 34.5\% of imperfect cases therefore contribute meaningful partial credit, and a back-of-envelope check confirms the arithmetic: if the 65.5\% perfect cases score 1.0 and the remaining 34.5\% score on average 0.71, the weighted average yields $(0.655$ x $1.0) + (0.345$ x $0.71) \approx 0.90$, consistent with our reported MIR-E.

The stage-by-stage breakdown reveals where errors enter the pipeline. The drop from heirs-correct (84.0\%) to all-correct (65.5\%) accumulates across three transitions: heir identification to share assignment ($-2.8$pp), share assignment to adjustment ($-1.8$pp), and adjustment to final distribution ($-13.9$pp). The largest single drop is the last, indicating that arithmetic computation in \texttt{post\_tasil} is the dominant bottleneck for cases that pass all upstream reasoning stages---consistent with calculation errors being the most impactful error category (5.8\% of cases, $-1.7$pp, Table~\ref{tab:error_dist}).

\subsection{Implications for Deployment}

Our error analysis reveals that the model is suitable for educational tools, preliminary screening, and simple cases, but requires human review for legal decisions, complex families ($\geq$8 relatives), rare adjustment cases, and high-stakes situations. Although the model scored 90.0\% MIR-E overall, the 10\% error rate remains too high for binding legal decisions without expert verification.

An optimal deployment strategy employs a hybrid human-AI workflow: (1) model generates initial analysis with 90\% accuracy, (2) automatic flagging of complex cases, rare cases ---\textAR{رد} (\textit{radd}) and \textAR{عول} (\textit{'awl})---, and cases where the model writes explicit fractions for known residuary heirs, (3) human expert reviews flagged cases, (4) model provides confidence scores to guide review priority. This approach leverages the model's strengths (speed, coverage of standard cases) while mitigating its weaknesses through targeted human oversight.

\section{Conclusion}

We presented QU-NLP's approach to the QIAS 2026 shared task on Arabic Islamic inheritance reasoning, achieving 90.0\% MIR-E through multi-stage QLoRA fine-tuning of Qwen3-4B followed by lightweight post-processing.
Our comprehensive error analysis reveals that:

\begin{enumerate}

\item The model demonstrates strong legal reasoning---97.1\% heir and blocking accuracy, 94.3\% share assignment accuracy---but exhibits specific weaknesses in adjustment detection (84.6\%) and final distribution arithmetic (5.8\% of cases show correct reasoning but wrong numerical output).

\item Per-category analysis shows \textit{'awl} cases (79.2\% MIR-E) underperform \textit{radd} cases (83.0\%) despite nearly eight times more training examples, revealing arithmetic complexity rather than data volume as the primary bottleneck. \textit{Radd} cases achieve 100\% heir and share accuracy with failures confined to the final arithmetic output step---evidence of generalised legal reasoning rather than pattern memorisation.

\item The PostTasil post-processing variant produces results identical to Basic across all metrics, confirming that the model's adjustment arithmetic is correct in every case it correctly classifies the adjustment type. The 29 remaining calculation errors are irreducible text-generation failures in the final output field, not addressable by any post-hoc symbolic layer.

\item Residue label recall is 24.7\% (103/417 cases requiring the label), with the model substituting explicit fractions in 75.3\% of cases. The global MIR-E cost is only $-0.85$pp because 83.1\% of avoidance cases compute the numerically correct residue fraction---confirming the failure is representational (wrong output format) rather than computational (wrong arithmetic), and is directly addressable through constrained decoding without retraining.

\item Domain-specific pre-adaptation on Islamic legal texts improves structured output quality, with fatwa records providing exposure to jurisprudential terminology and reasoning patterns not present in general pre-training data.

\end{enumerate}

Our results demonstrate that small fine-tuned models (4B parameters)
can match or exceed larger general-purpose models (32--120B parameters)
on specialized legal reasoning tasks, achieving commercial-grade
performance with minimal computational resources suitable for
deployment on consumer hardware.

Future work will explore: (1) constrained decoding to guarantee valid \texttt{JSON} structure, correct arithmetic, and enforcement of \textAR{باقى التركة} when \textAR{عصبة} heirs are present and fixed shares sum to less than unity---directly addressing the 75.3\% residue avoidance rate ($-0.85$pp) without retraining; (2) reinforcement learning from process rewards to improve multi-step reasoning; (3) oversampling rare adjustment cases---\textAR{رد} (\textit{radd}, 2.8\%) and \textAR{عول} (\textit{'awl}, 4.9\%)---to address training imbalance; (4) extension to other Islamic legal domains (marriage, divorce, financial transactions); and (5) hybrid systems combining neural reasoning with symbolic rule engines for guaranteed correctness in safety-critical applications.

\bibliographystyle{lrec2026-natbib}
\bibliography{ref}

\end{document}